\documentclass[lettersize,journal]{IEEEtran}
\usepackage{mathtools}
\usepackage[hidelinks]{hyperref}

\usepackage{amsmath,amsfonts}
\usepackage{algorithmic}
\usepackage{algorithm}
\usepackage{array}
\usepackage[caption=false,font=normalsize,labelfont=sf,textfont=sf]{subfig}
\usepackage{textcomp}
\usepackage{stfloats}
\usepackage{url}
\usepackage{verbatim}
\usepackage{lipsum}
\usepackage{graphicx}
\usepackage{cite}
\begin{document}

\title{ParaColorizer: Realistic Image Colorization using Parallel Generative Networks}

\author{Himanshu Kumar*, Abeer Banerjee*, Sumeet Saurav, and Sanjay Singh

\thanks{*These authors contributed equally to this work.}
\thanks{The authors are with the Intelligent Systems Group of Central Electronics Engineering Research Institute (CSIR-CEERI) Pilani, Rajasthan. (e-mail:himanshu.ceeri20a@acsir.res.in, abeer.ceeri20a@acsir.res.in, sumeet@ceeri.res.in, sanjay@ceeri.res.in)}}

\markboth{PREPRINT - ParaColorizer: Realistic Image Colorization using Parallel Generative Networks}%
{Shell \MakeLowercase{\textit{et al.}}: A Sample Article Using IEEEtran.cls for IEEE Journals}

\maketitle

\begin{abstract}
Grayscale image colorization is a fascinating application of AI for information restoration. The inherently ill-posed nature of the problem makes it even more challenging since the outputs could be multi-modal. The learning-based methods currently in use produce acceptable results for straightforward cases but usually fail to restore the contextual information in the absence of clear figure-ground separation. Also, the images suffer from color bleeding and desaturated backgrounds since a single model trained on full image features is insufficient for learning the diverse data modes. To address these issues, we present a parallel GAN-based colorization framework. In our approach, each separately tailored GAN pipeline colorizes the foreground (using object-level features) or the background (using full-image features). The foreground pipeline employs a Residual-UNet with self-attention as its generator trained using the full-image features and the corresponding object-level features from the COCO dataset. The background pipeline relies on full-image features and additional training examples from the Places dataset. We design a DenseFuse-based fusion network to obtain the final colorized image by feature-based fusion of the parallelly generated outputs. We show the shortcomings of the non-perceptual evaluation metrics commonly used to assess multi-modal problems like image colorization and perform extensive performance evaluation of our framework using multiple perceptual metrics. Our approach outperforms most of the existing learning-based methods and produces results comparable to the state-of-the-art. Further, we performed a runtime analysis and obtained an average inference time of 24ms per image. 
\end{abstract}

\begin{IEEEkeywords}
Information Restoration, Realistic Colorization, Parallel GANs, ResUNet, Image Fusion
\end{IEEEkeywords}

\section{Introduction}
\begin{figure}[h]
\centering
\includegraphics[width=0.485\textwidth]{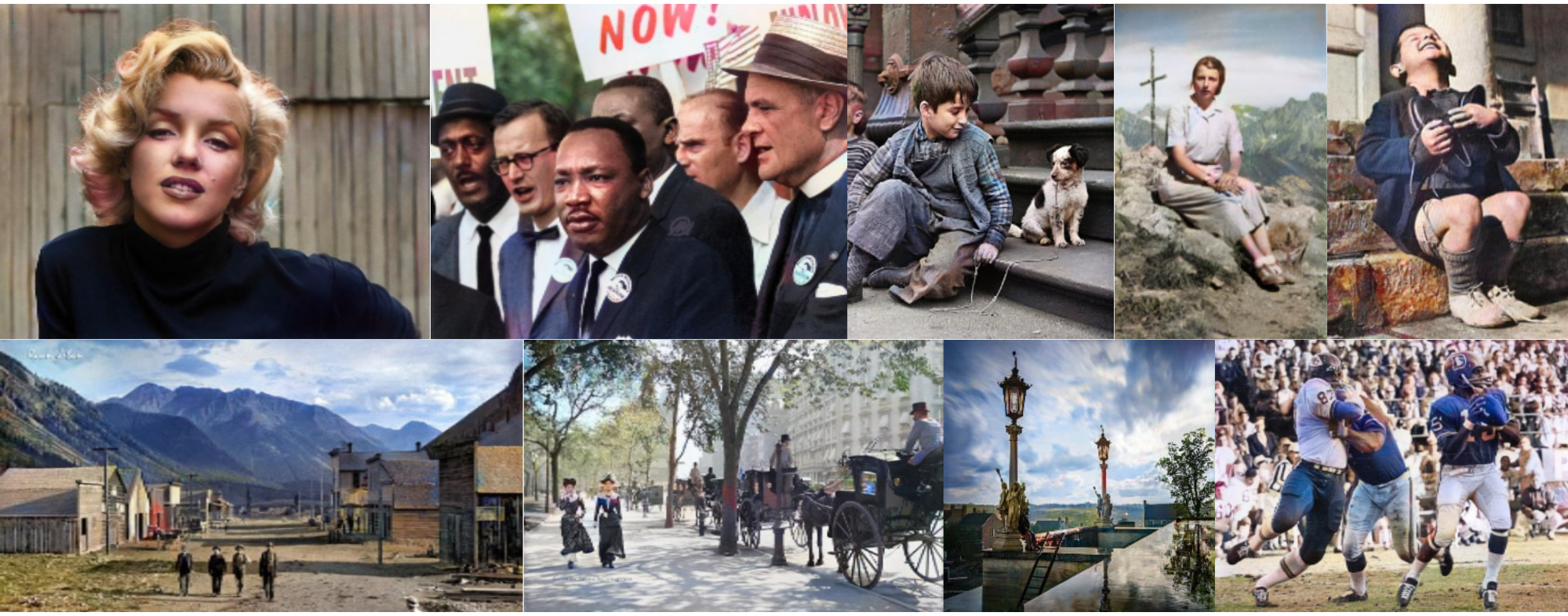}
\caption{Samples of colorized outputs using our method. The outputs are displayed as obtained without any further enhancements. Image courtesy: \url{shorturl.at/ckpK0} and \url{shorturl.at/lzY29}.}
\label{tag}
\end{figure}

\IEEEPARstart{G}{rayscale} image colorization is a well-known problem in the field of computer vision. Due to the fundamentally ill-posed nature of this problem, the solutions can be multimodal~\cite{charpiat2008automatic}, i.e., there might not be a particular solution for this task. There can be multiple plausible output images for a single input grayscale image (e.g., a hat can be red, blue, green, etc.). Potential applications of colorization can range from reviving the authentic significance from historical images and colorizing old black-and-white movies, to information compression. Although there have been multiple approaches for user-guided image colorization, fully automatic image colorization still remains a challenging research problem. Few of the results achieved using our approach are shown in Figure \ref{tag}. 

Earlier work in this field used a human-level interpretation-guided method ~\cite{levin2004colorization,huang2005adaptive,10.1109/TIP.2005.864231,10.1145/1141911.1142017,10.5555/2383847.2383887,Skora2009LazyBrushFP} to colorize the whole image. Primarily scribbles are applied on a grayscale image as baseline structured and the whole image is colorized from it. Using scribbles over an image is not a convenient task and it requires guidance from expert artists. In the advancement of guided methods, the reference-image based method~\cite{10.1145/566654.566576,ironi2005colorization,charpiat2008automatic,gupta2012image,liu2008intrinsic,gatys2016image} is being used which removes the dependency of expert artists by providing the reference image as a baseline idea to colorize the whole image. Recent development in deep learning~\cite{mirza2014conditional} accelerates the predictive modelling-based image colorization by generating the two missing channels. Generally, an image consists of a foreground and a background, where the foreground consists of object(s) of interest, and the background is the remaining image. Earlier deep-learning-based methods~\cite{gatys2016image,iizuka2016let,larsson2016learning,10.1007/978-3-319-46487-9_40,https://doi.org/10.48550/arxiv.1705.02999,isola2017image,https://doi.org/10.48550/arxiv.1808.01597,he2018deep,guadarrama2017pixcolor,BMVC2017_85,deshpande2017learning,Messaoud_2018_ECCV,antic2019jantic} performed learning and colorization on the entire image which usually led to poor performance in images with multiple objects but without clear figure-ground separation. This problem was addressed by instance-aware colorization~\cite{Su2020InstanceAwareIC}. Although this method colorizes the objects properly in most cases, in some cases, the context is missed and the background is left desaturated because a single model for colorization of the whole image is insufficient at covering the diverse data modes.

\begin{figure*}[t]
\centering
\includegraphics[width=1.0\textwidth]{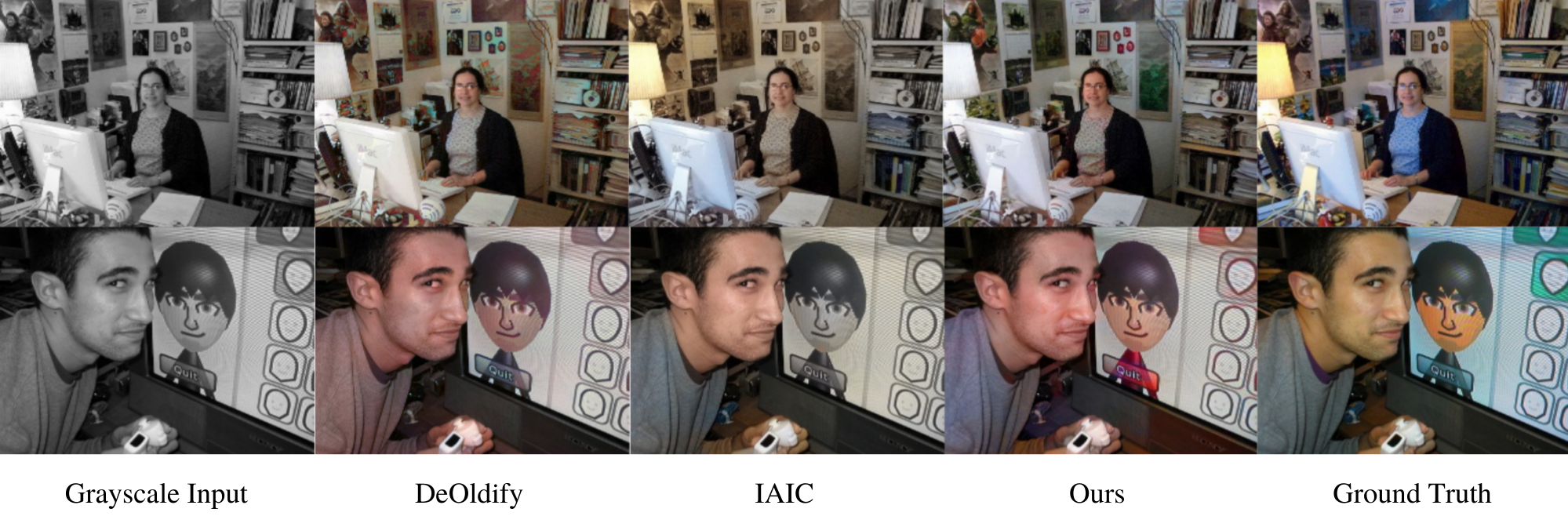}
\caption{The selected images have cluttered/confusing backgrounds which make the colorization task difficult even for existing state-of-the-art methods. DeOldify~\cite{antic2019jantic} suffered from color bleeding and IAIC~\cite{Su2020InstanceAwareIC} had desaturated background and both frameworks failed to capture the data diversity.}
\label{tagcompare}
\end{figure*}

In this paper, we aim at the automatic colorization of grayscale images while preserving contextual information using a novel deep-learning-based adversarial training framework. Unlike the existing SOTA method~\cite{Su2020InstanceAwareIC} that uses cropped objects from a large-scale common objects dataset to fine-tune a pre-trained model on full images to obtain the object-level semantic information, we deployed two parallel networks to colorize foreground and background separately and an adaptive fusion network which results in the final colorized output. Our main intuition was to employ multiple generators, instead of using a single one which has been proven to be extremely effective at covering diverse data modes while the whole image information is preserved~\cite{Hoang2018MGANTG}. Our whole colorization framework consists of the following salient steps. First, inspired by the unbalanced GAN approach, we pre-train the generators of both the GAN pipelines before the adversarial training to prevent the discriminator from winning too easily~\cite{Ham2020UnbalancedGP}. Second, adversarial training of a foreground network using an off-the-shelf pre-trained object detector to detect the potential objects corresponding to the images from the main dataset. Parallel training of a background network for which, common real-world scene-based examples are appended to the main dataset.  Third, a deep-learning-based fusion network inspired by DenseFuse~\cite{8580578}, trained on the main dataset, is deployed which adaptively merges the generated outputs of the foreground and background network to generate the final colorized image.
Our main contributions are:
\begin{itemize}
    \item A parallel adversarial training framework for edge-preserved image colorization using Unbalanced GANs.
    \item Separately reconfigured Residual-UNets specializing in foreground and background colorization which preserves contextual information.
    \item Design of a DenseFuse-based fusion network for feature-based adaptive fusion of the generated parallel outputs.
    \item Achieving realistic colorization results comparable to state-of-the-art with substantially lesser training data.     
\end{itemize}

The remaining contents of the paper are organized as follows: Section \ref{relwork} provides a quick overview of some of the most prominent ways for image colorization using various methods like user-guided, example-based, and learning-based image colorization. Section \ref{method} elaborates our motivation and methods for achieving realistic colorization. Section \ref{experiments} discusses the different types of experiments performed to train a model that can produce results close to the state-of-the-art. This section also discusses the fundamental problems with the existing evaluation frameworks for image colorization models. Section \ref{ablation} provides an extensive ablation study of the individual models, the loss functions used, and the fusion model. Section \ref{runtime} provides an analysis of the average inference time of our model and the computational resources used to reproduce the colorization results. Section \ref{discussion} discusses the strengths and concerns of our method and states the failure cases of our model. Finally, in Section \ref{conclusion}, we conclude this paper by highlighting our key observations and we further discuss the direction of future research on image colorization.

\section{Related Works}
\label{relwork} 
\subsection{User-guided colorization}

These methods~\cite{levin2004colorization,huang2005adaptive,10.1109/TIP.2005.864231,10.1145/1141911.1142017,10.5555/2383847.2383887,Hoang2018MGANTG} use human interaction to assign the initial strokes or color points on a grayscale image that become the initial parameters for colorizing the whole image. A method was proposed by Levin et al.~\cite{levin2004colorization} in which the nearby pixels accounted for the same color based on the similar luminance by solving a Markov random field to propagate scribble colors. Scribble-based work continues with the advancement in similarity metrics, for instance, An et al.~\cite{10.1145/1360612.1360639} proposed an energy-optimization method in which the initially coarse edits propagate to refined edits based on the policy that similar edits are applied to spatially-close regions. Some methods~\cite{huang2005adaptive} involve edge-detection to prevent color bleeding problems or color propagation based on color texture similarity~\cite{10.1145/1141911.1142017,10.5555/2383847.2383887}. The hybrid method proposed by Zhang et al.~\cite{https://doi.org/10.48550/arxiv.1705.02999} in which both deep learning and scribble are used that drastically reduced the human effort just by assigning the color hints. 

\subsection{Example-based colorization}

Unlike the previous works, they used reference image-guided methods to colorize grayscale images. The guided methods~\cite{10.1145/566654.566576,ironi2005colorization,charpiat2008automatic,gupta2012image,liu2008intrinsic,gatys2016image,he2018deep} are broadly divided into global and local transfer methods. In the global transfer method~\cite{5540201,1544887,10.1145/1141911.1142017}, the targeted result is matched with the reference image using the global features like mean, variance and histogram. Since these methods missed the low-level semantic pixel information the generated output seems unrealistic. Local transfer methods~\cite{10.1145/2601097.2601101,10.1145/2601097.2601137,wu2013content} on the other hand, preserves the low-level information at pixel level~\cite{10.1145/566654.566576,liu2008intrinsic}, super-pixel level~\cite{gupta2012image,guadarrama2017pixcolor} and semantic segments level~\cite{ironi2005colorization,charpiat2008automatic}. Example-based methods rely on the quality of the reference image that includes manual annotation of image region~\cite{ironi2005colorization,gatys2016image} which is a tedious process. To address this issue, learning-based methods are used where the model learns the mapping from a large dataset~\cite{he2018deep} to colorize the image and the video~\cite{https://doi.org/10.48550/arxiv.1906.09909}.

\subsection{Learning-based colorization}

These methods~\cite{gupta2012image,cheng2015deep,iizuka2016let,larsson2016learning,10.1007/978-3-319-46487-9_40,https://doi.org/10.48550/arxiv.1705.02999,isola2017image,https://doi.org/10.48550/arxiv.1808.01597} are fully autonomous in which the whole image is generated from the pixel level. Deep learning algorithm used as the backbone in which end-to-end learning involves a large-scale dataset such as e.g., ImageNet~\cite{5206848}. To colorize a grayscale image, broadly two kinds of problems are addressed by the recently proposed method: semantics and multi-modality~\cite{charpiat2008automatic}. The semantics-based method proposed by Iizuka et al.~\cite{iizuka2016let} and Zhao et al.~\cite{https://doi.org/10.48550/arxiv.1808.01597} uses two branched architectures to learn local image features and global priors and fuse them to colorize the complete image. Further, Zhang et al~\cite{10.1007/978-3-319-46487-9_40} proposed a cross-channel encoding scheme to preserve the semantics. Larsson et al.~\cite{larsson2016learning} also achieved semantic interpretability by pre-training the model for a classification task. To address the multi-modality of the problem, recent works~\cite{larsson2016learning,10.1007/978-3-319-46487-9_40,https://doi.org/10.48550/arxiv.1808.01597} proposed color distribution at per pixel level and achieved plausible results in most cases but missed the object level semantics.

To address this problem, Su et al. (IAIC)~\cite{Su2020InstanceAwareIC} proposed a method in which an object detector is used for instance-based fine-tuning of the baseline. Although the instances are colorized properly, the backgrounds are left desaturated and the excellence of the model depends on the object detector’s performance. We observed that obtaining the semantic information at the object level while maintaining the foreground and background context is difficult to achieve using a single deep-learning network. Thus, we employ parallel networks specifically tailored at maintaining the foreground semantics and the background scene separately.

\subsection{Image Colorization with GANs} Attempts have been made to colorize grayscale images using GAN-based frameworks \cite{nazeri2018image}. Vitoria et al. \cite{vitoria2020chromagan} proposed a self-supervised adversarial model that performs colorization by using the perceptual and semantic understanding of color, while DeOldify \cite{antic2019jantic} uses progressive GANs for colorization of historical images. Both of these methods have been included for comparison with our approach. Style transfer-based anime sketch colorization ~\cite{zhang2017style} has been performed using a Residual-UNet-based generator. Parallel networks have been previously used by Johari et al. \cite{o2009context} for generating images with similar color themes. Our method is unique and advanced in several manners. First, we parallelly train two different GAN architectures to cover the data diversity. \textit{(a) Foreground pipeline:} Consists of a Residual-UNet based generator with self-attention and is trained using an edge-aware loss function. \textit{(b) Background pipeline:} Works as a compensator to the foreground model and is separately modified for colorizing common backgrounds. Secondly, using an unbalanced training method \cite{Ham2020UnbalancedGP} by pre-training the generator to prevent the early convergence of the discriminator. Thirdly, using a DenseFuse-inspired fusion network to obtain a final fused image that performs the feature-based adaptive fusion of the parallelly generated outputs. 

\begin{figure*}[t]
\centering
\includegraphics[width=1.0\textwidth]{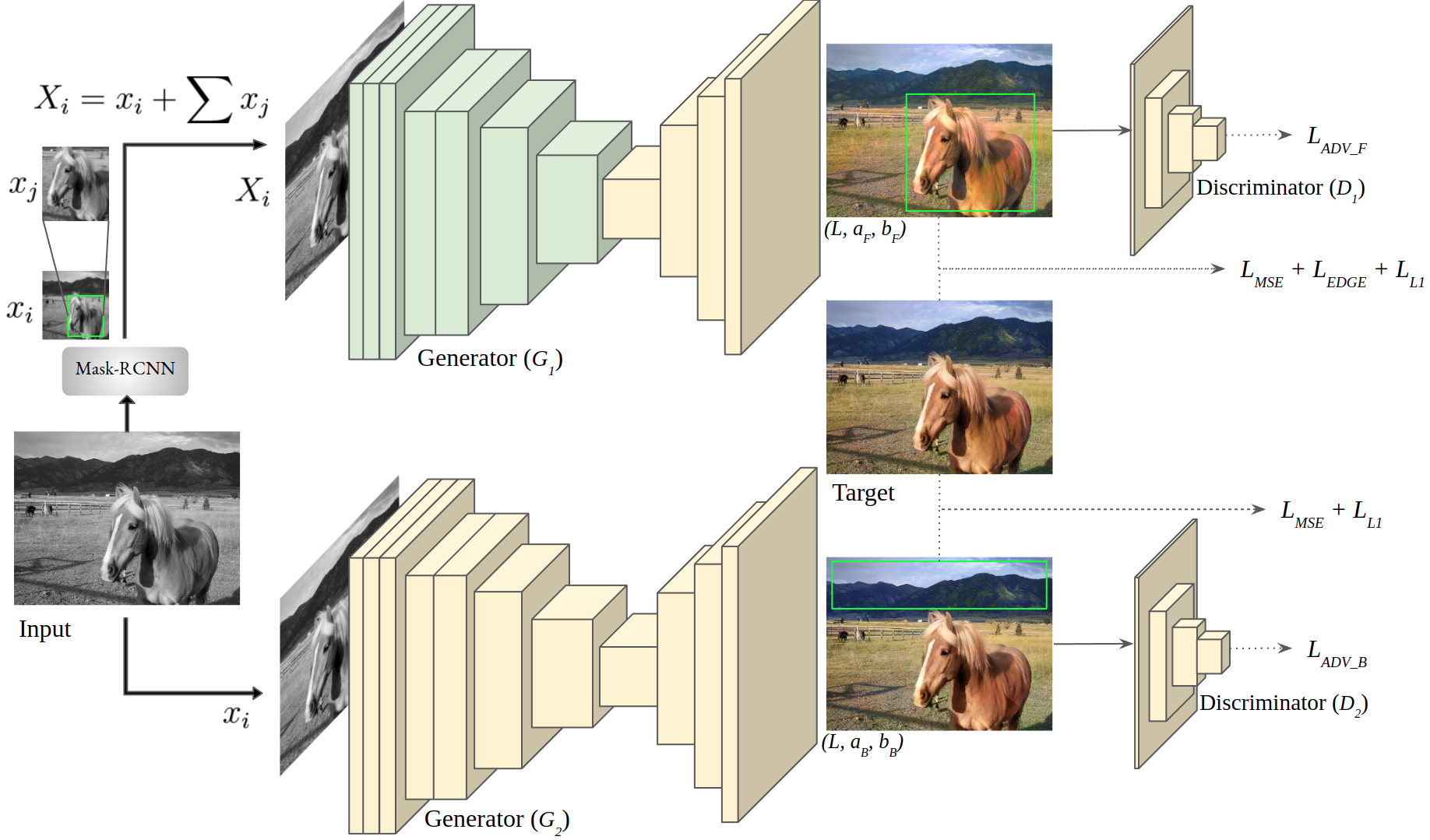}
\caption{The adversarial training pipeline. Generator $G_1$ and generator $G_2$ are Res-UNets (Generator $G_1$ has self-attention enabled) that accept L-channel at the input and predict the $(a,b)$ channels. The L-channel is combined with the predicted $(a,b)$ channels and passed to PatchGAN discriminators $D_1$ and $D_2$ in each forward pass. $L_{MSE}$, $L_{EDGE}$, $L_{L1}$, $L_{ADV}$ denote the mean-squared loss, the edge-map loss, the L1 loss, and the adversarial loss respectively.}
\label{paratrain}
\end{figure*}

\section{Method}
\label{method}
The task of image colorization has been divided into two subtasks of foreground and background colorization. The basic idea is to employ two independently trained separately modified specialized models which parallelly aim at colorizing the input grayscale image and the results are fused by a fusion network to obtain the final colorized output. The complete training pipeline is shown in Figure \ref{paratrain}.

\subsection{Foreground Colorization}
A subtask that mainly aims at effectively colorizing the objects in the image is achieved by training the network on the full image along with their corresponding object proposals. The main objective of doing this is to simulate object attention which makes the network specialize at colorizing the foreground. In order to simulate object attention, the object proposals were obtained from the full images in the main dataset by making use of an off-the-shelf pre-trained network, Mask R-CNN~\cite{he2017mask}, as done in IAIC~\cite{Su2020InstanceAwareIC}. The full images $x_i$ were passed through the MaskRCNN framework with ResNet50 backbone to detect the object bounding boxes and the corresponding cropped object proposals denoted by $x_j$ were obtained.\\

\textit{Colorization Backbone Architecture:} We use a GANbased framework for colorizing the foreground which is heavily inspired by the Pix2Pix framework~\cite{isola2017image}. The generator model employs a modified version of UNet presented in Pix2Pix. Our generator is a UNet that uses a pre-trained ResNet-18 network as its encoder for simplifying the training process. We have also enabled a self-attention mechanism to model the long-range dependencies across image regions~\cite{zhang2019self}. The discriminator network is a PatchGAN as proposed in Pix2Pix which classifies patches of the total image instead of classifying the whole image as real or fake which helps to capture the local style statistics.\\

\textit{Loss Function and Training:} We use a weighted combination of three loss functions for training the foreground colorizer network. Firstly, the standard mean-squared error loss function described by (enter equation) as described in Least-Squares GAN~\cite{mao2017least} for stable learning and better quality of output images. Secondly, the standard L1 loss function encourages less blurring of output~\cite{isola2017image}. Thirdly, we implement an edge-map loss inspired by Xia et al.~\cite{xia2021edge} to reduce color bleeding and blurs around the image edges. The edge map is computed using the Canny edge detector (using Otsu’s thresholding method). We compute the edge maps of the ground truth image and the foreground colorized image and the edge map loss is defined as the mean of the difference between these two edge maps. 

Consider $x$ as the input grayscale image, $z$ as the input noise for the generator, $y$ as the ground truth color image, $G$ as the generator, and $D$ as the discriminator. For paired image-to-image translation, $x,y$ pair denotes the paired domain of grayscale and color images. Then the loss for our conditional GAN is:

\begin{align}
\begin{aligned}
L_{ADV}(G,D) &= E_{x,y}[\log D(x,y)]\\
&\qquad + E_{x,y}[\log (1 - D(x,G(x,z)))]  
\end{aligned}
\end{align}

The edge-map loss function involves the Canny edge detector which is denoted by $C$.  The basic idea is to compute the difference between the edge maps of the ground truth image $C(y)$ and the colorized image $C(G(x))$. If the colorized output is suffering from color bleeding artifacts, the edge map of the colorized output will be different from the ground truth, thus the edge-loss will be more. The value of the edge loss will be less only if the edge maps come out to be similar to each other. So, the loss functions being used are:

\begin{equation*}
L_{EDGE}(x,y,G) = E_{x,y,z}[\mid \mid C(y) - C(G(x,z)) \mid \mid]   
\end{equation*}

\begin{equation*}
L_{L1}(G) = E_{x,y,z}[\mid \mid y - G(x,z) \mid \mid]   
\end{equation*}

Therefore, the final loss function is:
\begin{align}
\label{edgeq}
\begin{aligned}
G^* &= arg\:min_G\:max_D\:L_{ADV}(G,D)\\
&\qquad + \lambda_1 L_{EDGE}(G) + \lambda_2 L_{L1}(G)
\end{aligned}
\end{align}

\subsection{Background Colorization}
In this case, the complexity of the training pipeline is reduced because background colorization does not generally involve colorizing sharp edges since only the foreground stays in focus. The background colorized image is obtained by combining the predicted channels with the luminance channel $L$ to form $(L, a_B, b_B)$. Image datasets with more examples of common scenes, places, and landscapes can serve as better examples of background. Public datasets containing the same are appended to the main dataset. This time the network is not provided with object proposals.\\

\textit{Colorization Backbone Architecture:} The background colorizer network is simply a slightly modified version of the Pix2Pix framework. The generator model is a simple Residual UNet (with the same pre-trained ResNet-18 encoder) without self-attention. This is chosen to reduce the model complexity as colorizing backgrounds is a relatively easier task. The same PatchGAN discriminator is being used here as proposed in Pix2PIx.\\

\textit{Loss Function and Training:} A combination of MSE loss function and L1 loss function is used to optimize the background colorizer. This model should act as a compensator to the foreground colorizer model so the edge map loss is not significant in this case. 

The overall loss function for the background model is, therefore: 

\begin{equation*}
G^* = arg\:min_G\:max_D\:L_{ADV}(G,D) + \lambda L_{L1}(G)    
\end{equation*}

\subsection{Fusion Network}
The fusion network was inspired by the DenseFuse~\cite{8580578} framework that was used for the fusion of visible and infrared images. Here, we redesigned and repurposed the network to merge the outputs of the foreground and background colorizer models. It is to be noted that the foreground and background colorizer models are self-sufficient models at full colorization of an image and they are named such that their specialization area is identified. This network is required at the testing pipeline for the adaptive fusion of the colorized foreground-background pair, so this can be trained independently. 

\textit{Training and Network Architecture:} The fusion network is a modified version of the DenseFuse~\cite{8580578} framework that was actually made for infrared thermal images. The network is modified to work with color images. It is an encoder-decoder framework with a three-layer dense convolutional block at the encoder. The network takes single color image as its input and is trained using a weighted combination of mean-squared error (MSE) loss and structural similarity index (SSIM) loss function because the aim of the network is to generate an identical copy of the input image. The training was performed on the COCO dataset~\cite{lin2014microsoft} for nine epochs.\\

\textit{Fusion Process:} In the testing pipeline, the pre-trained encoder of the fusion network receives a pair of three-channel colorized images and the features are computed. The features are added using feature fusion strategies that are employed in DenseFuse. The pre-trained decoder receives the combined features as its input and produces a three-channel color output as the final image.\\

\section{Experiments} 
\label{experiments}
To obtain the ideal behavior for image colorization using a parallel method, we conducted extensive experiments using different versions of generator architecture. The detailed experimental settings corresponding to each version have been discussed in the Ablation Study section. All the different versions of the parallel GAN framework had the same skeleton but were trained with different network settings to efficiently capture the contextual information of the image. We used the COCO and Places dataset to train the colorizer pipelines independently. We show significant counterexamples where the metric of PSNR fails to match human visual assessment, so we present a quantitative comparison report against state-of-the-art colorization models using multiple perceptual and non-perceptual evaluation techniques.

\subsection{Training Details} Our whole training procedure is divided into three stages. The first two stages correspond to the parallel adversarial training of the colorizer frameworks. The third stage is for training the fusion framework. The adopted training methodologies for training the colorizer models include the pre-training of generators inspired by the unbalanced GANs~\cite{Ham2020UnbalancedGP} framework. For the final model, we used generator and discriminator learning rates of $2e-4$ with the Adam optimizer ($\beta_1 = 0.5, \beta_2 = 0.999$). Using the pre-trained generator, the adversarial training had to be performed for 15 epochs to obtain stable outputs. The fusion network was trained for 20 epochs with a learning rate of $1e-4$ using the Adam optimizer ($\beta_1 = 0.9, \beta_2 = 0.999$). All the training was performed on two 16GB NVIDIA Tesla V100 GPU nodes.\\

\textit{Pre-training the Generators:} The unbalanced GANs~\cite{Ham2020UnbalancedGP} framework showed faster convergence of the generator and the discriminator with a better image quality at much earlier epochs compared to ordinary GANs. The main objective of this step was to prevent the discriminator from winning too easily during the adversarial training and to maintain a stabilized learning process. We found that pre-training the generator even for just a few epochs can be a good start for the adversarial training process. The generators of the final version of the parallel GAN framework were pre-trained using the COCO Training Split dataset with a learning rate of $1e-4$ for 15 epochs using the Adam optimizer ($\beta_1 = 0.9, \beta_2 = 0.999$).

\subsection{Evaluation Metrics}
\label{evalmetrics}
\begin{figure}[h]
\centering
\includegraphics[width=0.485\textwidth]{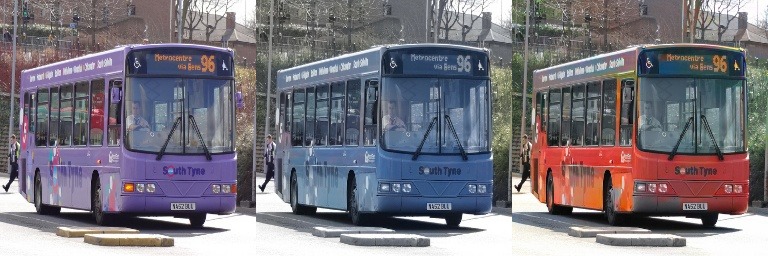}
\caption{The ground truth (left) image is recolorized using IAIC~\cite{Su2020InstanceAwareIC} (middle) and
our method (right). The PSNR of the middle image is 23.93 and the PSNR obtained using our method (left) is 16.80}
\label{psnr}
\end{figure}

Since plausible colorization results probably diverge a lot from the original color image, pixel-wise evaluation metrics like PSNR fail to model the human visual assessment. An image elaborating the issue is shown in Figure \ref{psnr}.
The first image being the ground truth, the second image is having a much higher PSNR compared to the third even though one can observe that the third image is better colorized with pronounced color details in both object and background. But, since PSNR does not account for multimodality, the color accuracy becomes the only determining factor. 
\begin{figure}[h]
\centering
\includegraphics[width=0.485\textwidth]{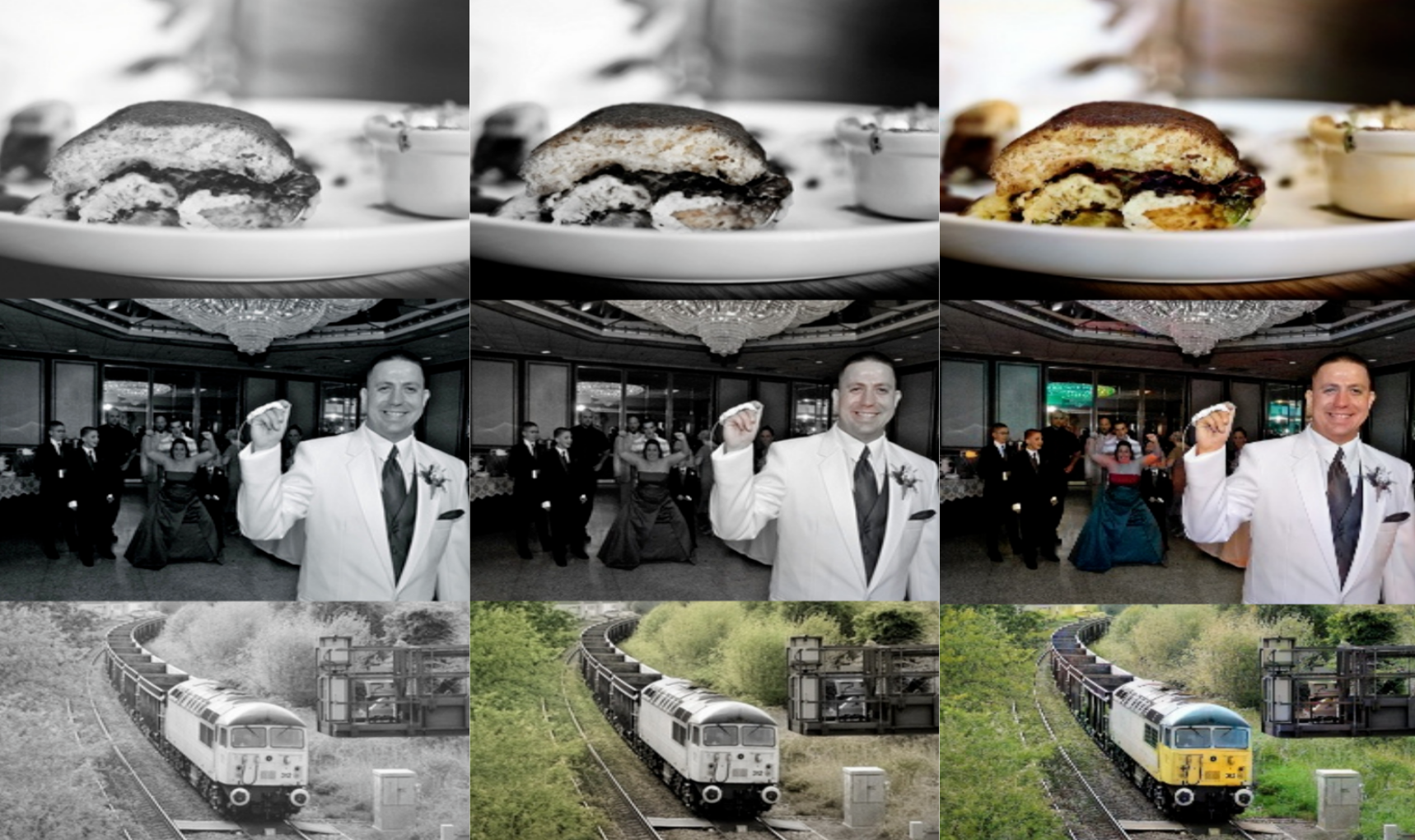}
\caption{Examples where the PSNR metric fails to identify a better colorized output. The middle column has a better PSNR compared to the right column.}
\label{psnr_comp}
\end{figure}

The dependence of PSNR on the ground truth makes the metric a good choice for producing accurate colors but for tasks like image restoration, even if the ground truth itself is corrupted, PSNR compels the model to produce a corrupted image that is similar to the ground truth that generates a higher PSNR as shown in the figure \ref{psnr_comp}. That's why perceptual metrics like FID, LPIPS, etc. are better suited for the quantitative comparison of image quality and fidelity. FID, instead of comparing pixel by pixel, compares the mean and standard deviation of one of the deeper layers in Inception model. We computed the FID using the official implementation of FID by PIQ~\cite{seitzer2020pytorch} using 5000 images of COCO validation split dataset resized to resolution 256x256. Inspired by Wu et al. \cite{wu2021towards}, we use Colorfulness score \cite{hasler2003measuring} to evaluate the vividness of the colorized outputs.

\subsection{Comparison Report}
We used the COCO validation split dataset with 5000 images resized to 256*256 for the quantitative comparison. While the state-of-the-art models mainly used large-scale datasets like ImageNet with 1.3M images for training, we used a much more compact and manageable COCO Training Split dataset and scene samples from the Places dataset \cite{zhou2016places}. Even with a substantially lesser amount of training data, our model was able to compete and achieve quantitative scores close to state-of-the-art methods. We evaluated our results using perceptual metrics like FID and LPIPS which are well-suited for evaluating the performance of generator outputs. FID specifically compares the similarity of the generated data distribution with that of the real data distribution making it a significant evaluation metric for judging the realism of the generated data. We also evaluate the PSNR and SSIM values for reference but it has been previously established that these metrics are not reliable for evaluating colorization results \cite{wu2021towards}. The results are discussed in Table \ref{cocotable}.
\begin{table}[h]
\centering
\begin{tabular}{ llllll }  
\hline
\\
Method & $\Delta$Colorful$\downarrow$ & FID$\downarrow$ & LPIPS$\downarrow$ & PSNR$\uparrow$ & SSIM$\uparrow$\\    
\\
\hline
CIC~\cite{10.1007/978-3-319-46487-9_40}                  & 23.40     & 33.73     & 0.234     & 21.83     & 0.895 \\
LRAC.~\cite{larsson2016learning}                & 31.57     & 31.66     & 0.183     & 25.06     & 0.930 \\
DeOldify~\cite{antic2019jantic}             & 8.12      & 23.09     & 0.180     & 23.69     & 0.920 \\
ChromaGAN~\cite{vitoria2020chromagan}            & 5.84      & 30.51     & 0.159     & 23.60     & 0.911 \\
IAIC~\cite{Su2020InstanceAwareIC}                 & 9.60      & 17.66     & 0.125     & 27.77     & 0.940 \\
\bf{Ours}                         & \bf4.90   & \bf17.60  & 0.151     & 23.96     & 0.925 \\
\hline
\\
\end{tabular}
\caption{Quantitative comparison of colorization methods. Our method
was trained on COCO (Training Split) and Places only.}
\label{cocotable}
\end{table}

In order to provide additional comparison results with ColTran \cite{kumar2021colorization}, a recent learning-based method who have used FID for evaluating their results, we randomly selected 5k images from the ImageNet Testing Split dataset as performed by ColTran, using which our model was able to achieve better FID scores compared to state-of-the-art models. (see Table \ref{imagenettable}).
\begin{table}[h]
\centering
\begin{tabular}{ llll } 
\hline
Method  & ColTran~\cite{kumar2021colorization}  & IAIC~\cite{Su2020InstanceAwareIC} & \textbf{Ours}\\    
\hline
FID$\downarrow$ & 22.0613 & 18.6942 & \textbf{16.8273}\\
\hline
\\
\end{tabular}
\caption{Comparison with state-of-the-art methods on ImageNet Testing Split dataset.}
\label{imagenettable}
\end{table}

\subsection{Visual Comparison} 
\begin{figure*}
\centering
\includegraphics[width=1.0\textwidth]{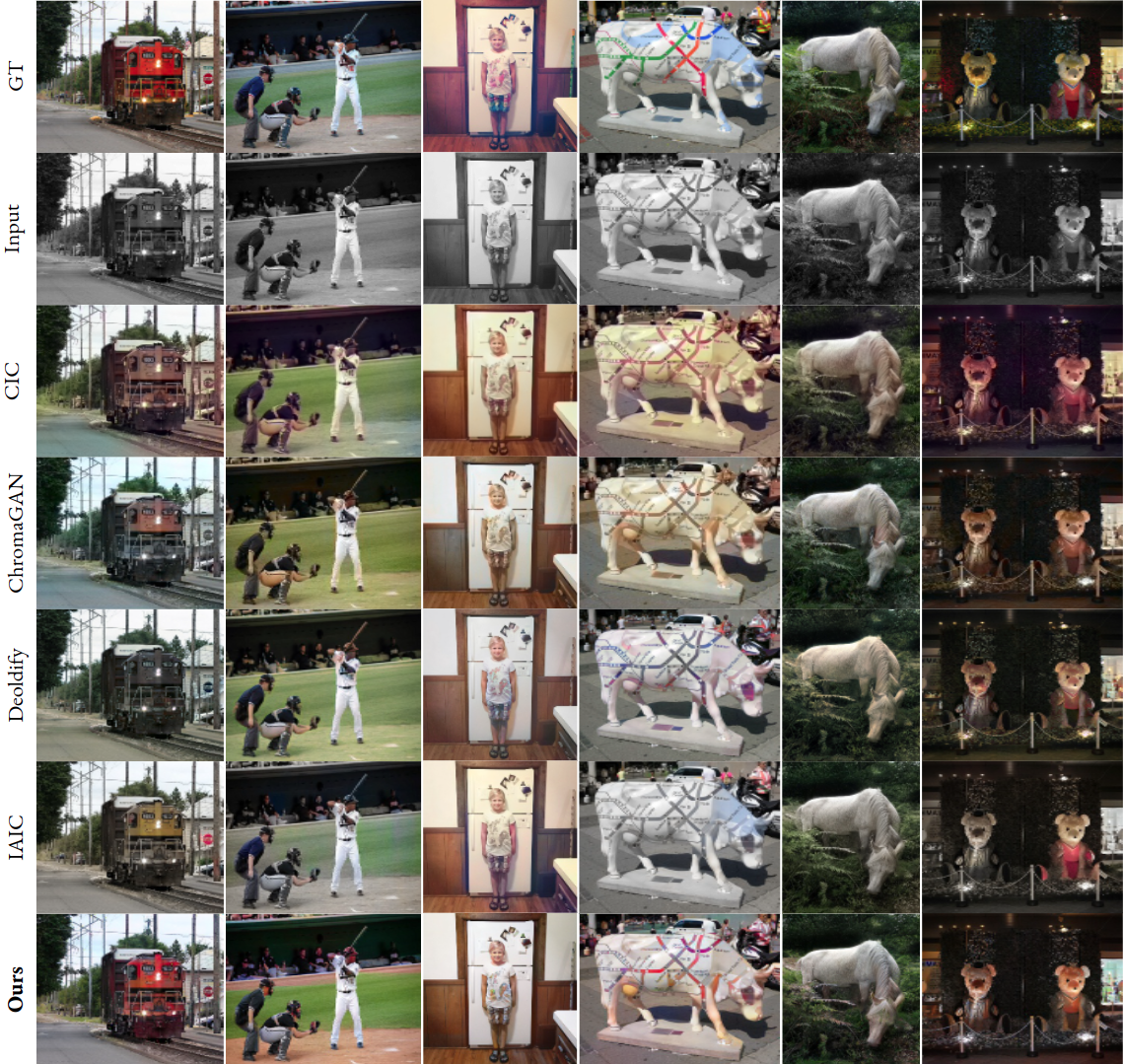}
\caption{This grid compiles the visual comparison results of existing methods and ours. Our method outperforms the existing methods by accurately colorizing the foreground and background while maintaining the edges and the long-range contextual dependencies. RTUG~\cite{https://doi.org/10.48550/arxiv.1705.02999} DeOldify~\cite{antic2019jantic} IAIC~\cite{Su2020InstanceAwareIC}}
\label{rescompare}
\end{figure*}
\begin{figure}[h]
\centering
\includegraphics[width=0.485\textwidth]{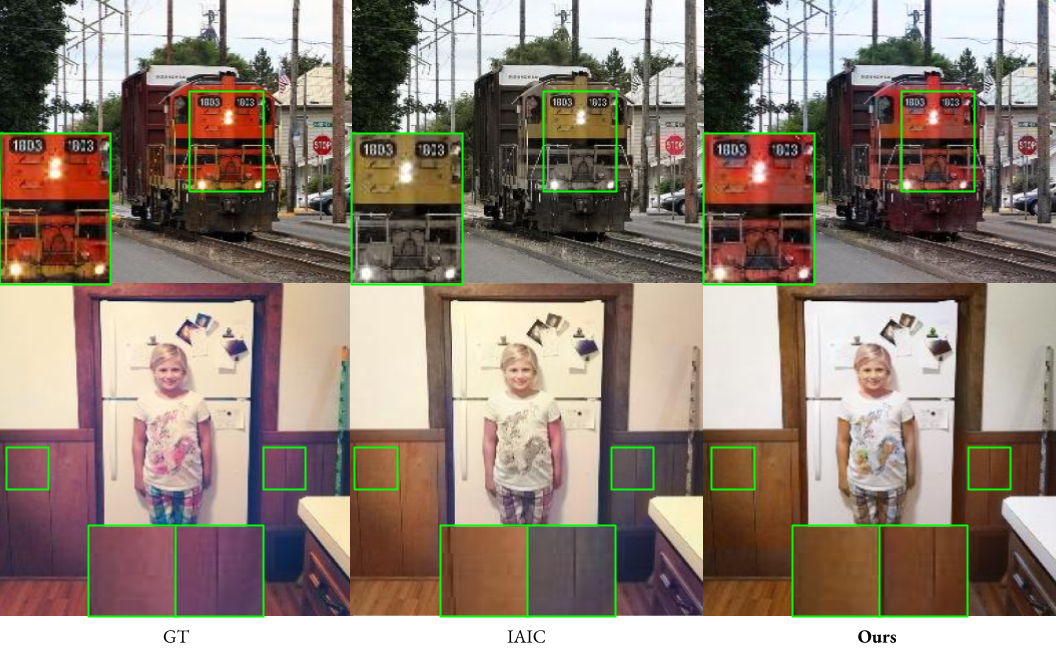}
\caption{This grid compiles the visual comparison results of the SOTA and ours. The first row shows the object has been completely colorized without any bleeding artifacts. The second row displays that our model could capture the long range spatial dependencies.}
\label{sotacompare}
\end{figure}

The visual comparison report is presented as a grid of images with a varied range of foreground-background combinations to illustrate the strength of our pipeline. The results are provided in Figure \ref{rescompare}. The first column consists of a dominant object in the foreground with a moderately cluttered background. To make the colorized images appear realistic, context preservation with accurate colorization of the subject in the frame is of utmost importance. The second column of images has a cluttered but prominent background (the stadium crowd) and a foreground with multiple objects (the baseball players). Our method is able to colorize the players, the stadium crowd  while colorizing the field. Other popular methods suffered context confusion (DeOldify) and desaturated backgrounds (IAIC). The third column requires the model to account for long-range spatial dependencies (achieved by using the self-attention mechanism) so that the left and the right side of the wooden door have the same color. Most methods fail to maintain the long-range color consistency, thus colorizing the left and the right side differently (IAIC). The fourth column has a prominent object with saturated colors. Our method is able to colorize the foreground while maintaining the edge details, whereas other methods showed a yellow bias (CIC, ChromaGAN). The fifth column represents natural images where most of the listed approaches performed flawlessly, the reason being clear figure-ground separation, whereas in the final column, even under a low illumination level, our model could colorize the complete image. An improved view of selected images is highlighted in Figure \ref{sotacompare}.

\subsubsection{Human Evaluation}
\begin{figure}[h]
\centering
\includegraphics[width=0.485\textwidth]{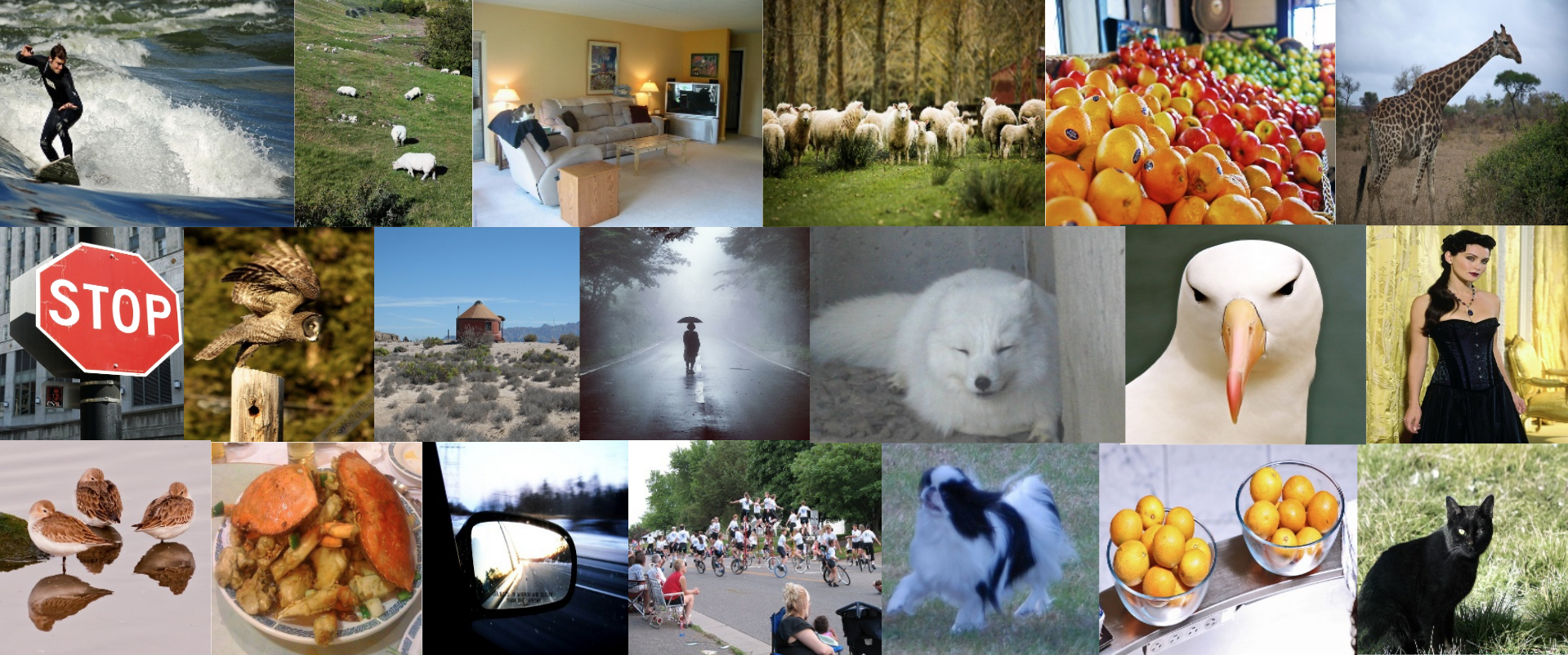}
\includegraphics[width=0.485\textwidth]{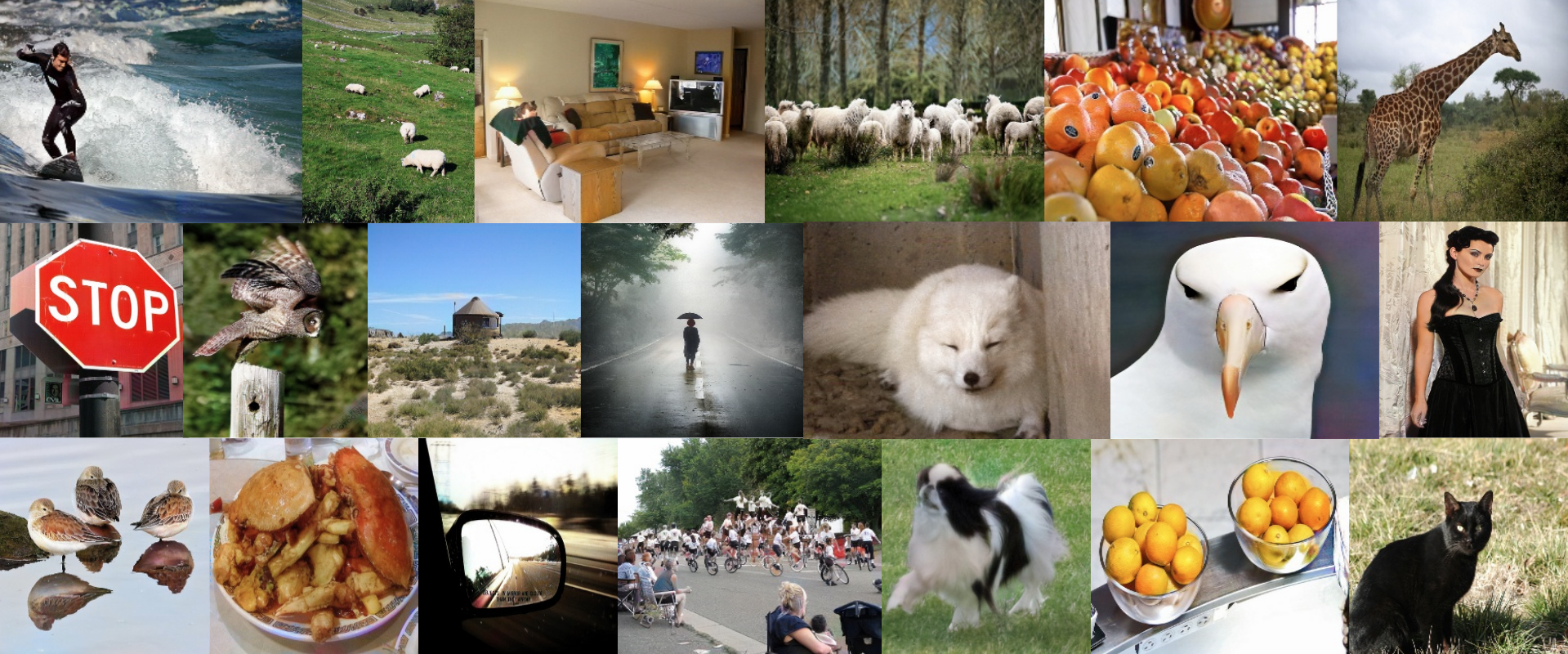}
\caption{The top section corresponds to the ground truth images and the bottom section is for the recolorized images. }
\label{realfake}
\end{figure}

To support our claims of realism, we performed human evaluation based the whether our colorized outputs looked real or not. Traditional human evaluation tests ask the subjects to rate the colorization quality of a given method. Whereas, we believe a blind photorealism evaluation of the generated images would provide a better estimate of whether a method can generate images that look natural. We prepared 4 batches of recolorized images, each with 5 pairs of recolorized images and their corresponding ground truths. The pairs were displayed to the test subjects and they had to identify the recolorized image generated by our colorization framework and provide a confidence score at the end of each batch. To make the subjects accustomed to the test, a trial run was prepared where the answers were shown to the subject. We chose 20 subjects from diverse educational backgrounds who were shown 20 images each, totalling 400 decisions. All the recolorized images and their corresponding ground truth images have been shown in Figure \ref{realfake}.

The results were obtained by performing the human evaluation test. The subjects had four options corresponding to each image that enabled them to classify the image as fake or real and to say whether they were confident about their prediction. The scoring strategy for the evaluation of our model was prepared on the basis of these four options that generated four possible cases after the subject has classified the image as real or fake: \textbf{Case 1}, the subject is confident about their prediction and they are right. Our score is 0. \textbf{Case 2}, the subject is not confident about their prediction but still, they are right. \textbf{Case 3}, the subject is not confident and is wrong. Here we can choose our scores to be 0.25 and 0.75 respectively. \textbf{Case 4}, the subject is confident but still is wrong. We get a score of 1.

For the first case, our model failed to fool the human subject, so it gets the minimum score and for the last case, our model was able to fool the subject completely i.e. it might have generated a better quality of colorized output compared to the real image, so it gets the maximum score. For the second case, the subjects were not sure but still were able to guess correctly, here, the model couldn't fool the subjects but at least produced results that could confuse the subjects, so it get 0.25. The third is when the model fools the subject but there is confusion, therefore we set the score as 0.75 since the model could deceive the subjects anyway. The fooling score $F$ was simply calculated from the table by the given equation where $V(I_i, C_i)$ is the entry corresponding to the $i$th row and $j$th column and $K_i$ is the score corresponding to the $i$th case; $N$ is the number of subjects participating in the test; $n$ is the maximum number of cases (=4 in this case); $b$ is the number of batches (=4 in this case)  
\begin{equation*}
F = \frac{\sum_{i=1}^{n}\sum_{j=1}^{b} V(I_i, C_j) * K_i}{b \times N}
\end{equation*}

The human evaluation score obtained by considering the mentioned scoring strategy is \textbf{0.5375}.

\begin{table}[h]
\centering
\begin{tabular}{lllll} 
\hline

Batch & Case 1 & Case 2 & Case 3 & Case 4 \\    

\hline
Batch 1     & 6     & 37    & 52     & 5 \\
Batch 2     & 12    & 40     & 40    & 8 \\ 
Batch 3     & 6     & 29    & 55     & 10 \\
Batch 4     & 7     & 34     & 49    & 10 \\
\hline
\end{tabular}
\caption{Table corresponding to the human evaluation test performed to evaluate the photorealism of the generated colorized images.}
\label{table}
\end{table}

\section{Ablation Study}
\label{ablation}
This section has been divided into three subsections where we perform an extensive ablation study of the individual models in the foreground and background colorization pipeline, the edge-loss function, and the fusion model. The foreground colorized output has a desaturated background but the objects (the donut, the woman, and her clothes) in the foreground were properly colorized, while the background colorized output compensates for the desaturated portions of the former by properly colorizing the complementary regions. The merged output finally combines the best of both the images to produce foreground-background colorized output.

\subsection{Ablation study of the Individual Models} 
\begin{figure}[h]
\centering
\includegraphics[width=0.49\textwidth]{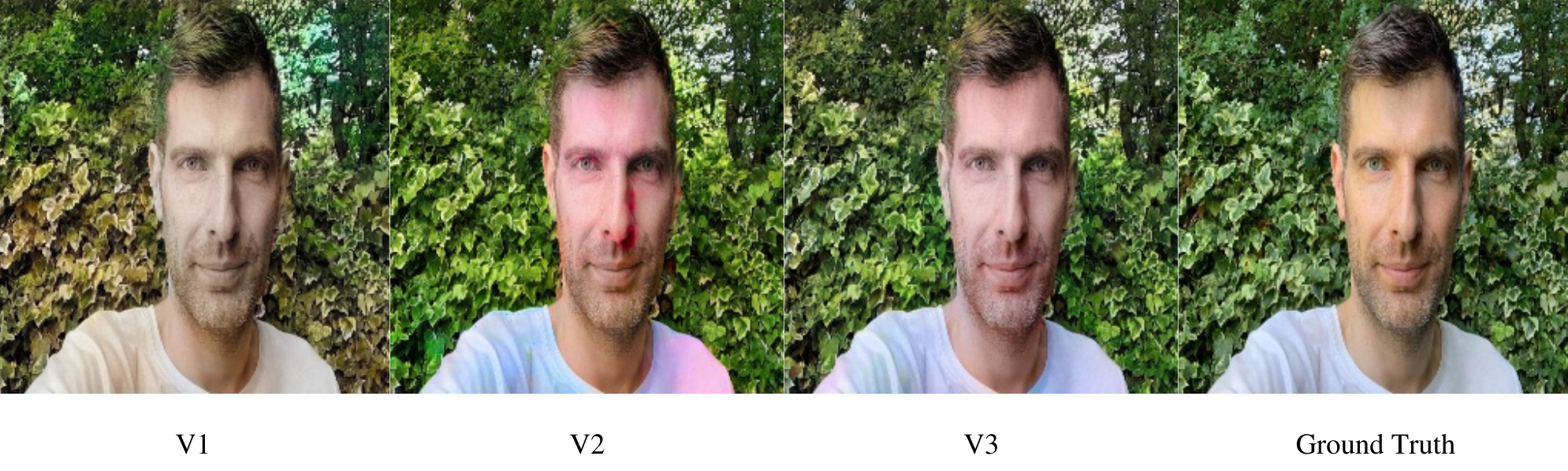}
\caption{Results of ablation study with different versions of colorizer frameworks.}
\label{versions}
\end{figure}

Different versions (shown in Figure \ref{versions}) of the pipeline are tested with different experimental settings to obtain generators specific to foreground and background colorization tasks. For the foreground colorizer pipeline, we find that adding self-attention to the generator Res-UNet improves the performance by throwing consistent colors across the spatial range of the image to account for long-range dependencies. Including the edge-map loss is also found to reduce the color-bleeding problem in the final colorized image thus increasing the FID score of the model. For the background colorizer, isolated performance decreases compared to the foreground colorizer.
\subsubsection{UNet Generator (V1)} We implemented a basic Pix2Pix based end-to-end pipeline for object and scene colorization independently where we used the COCO dataset and its cropped object proposals to train the UNet based generator for the foreground colorizer pipeline. The PatchGAN based discriminator was employed for adversarial training which could distinguish between the ground truth and the generated colorized output. The background colorizer pipeline was trained on full image features derived from the COCO dataset. Both pipelines used L1 loss for updating the generator. The model was trained for 15 epochs with generator and discriminator learning rates of 2e-4 using the Adam optimizer ( $\beta_1 = 0.5, \beta_2 = 0.999$).

\subsubsection{Res-UNet Generator (V2)} We used a ResNet-18 network as the encoder of the UNet based generator to capture the object information more effectively. To make the generators more robust, we used unbalanced training~\cite{Ham2020UnbalancedGP} in this version so that the discriminator does not win too easily over the generator. We are using the ResNet18 network as the encoder of the Residual-UNet architecture of both the pipelines and we train the generator only to condition it for a particular task. Other experimental settings were kept the same. Since the previous version was unable to colorize the scene-based examples, we appended Places dataset samples in the background colorization pipeline and it produced acceptable results.

\subsubsection{Self-Attention Res-UNet Generator (V3)} Deployment of unbalanced training with Res-UNet architecture drastically improved the convergence rate of the model and the addition of the samples of the Places dataset resulted in a better performance in complicated test cases. We further modified our model to incorporate the long-term dependency by introducing self-attention to the network due to which color artifacts were reduced and color consistency was maintained in the colorized outputs. 

We identified the problem of color bleeding in the previous versions so an edge loss function was incorporated with the L1 and MSE loss functions in the generator of the foreground colorizer pipeline. The final modified version addressed all the problems discussed in the above experimental settings and efficiently dealt with the diverse data modes and generated the foreground and background colorized proposals for the fusion network.

\subsection{Ablation Study of Edge Loss}
The edge loss ablation study has been performed to study the effect of edge loss on color bleeding. The factor $\lambda_1$ decides the weight of edge loss in the overall loss function given by the equation \ref{edgeq}.

Since color bleeding has adverse effects on the structural similarity between the ground truth and the colorized image, we chose the perceptual metric of Structural Similarity Index (SSIM) to evaluate the performance of our method with varying values of $\lambda_1$. Since the edge loss function was applied only to the foreground colorizer, the SSIM results have been obtained on the individual colorized outputs of the foreground colorizer. The training pipeline provided in Figure \ref{paratrain} was followed for each version of the model with different values of $\lambda_1$ and the results are discussed in Table \ref{lamda}. 

\begin{table}[h]
\centering
\begin{tabular}{ l|lllll } 
\hline
$\lambda_1$         & 0.00  & 0.25  & 0.50 & 0.75 & 1.00\\
SSIM $(\uparrow)$   & 0.863  & 0.894  & 0.900  & 0.915  & 0.924\\
\hline
\end{tabular}
\caption{Edge loss ablation study: Performed by varying $\lambda_1$ to obtain the SSIM results on the outputs of the foreground colorizer.}
\label{lamda}
\end{table}

\subsection{Ablation Study of Fusion Model}
The colorization performance achieved by the individual colorizer models has been computed to study the effect of the fusion model on the colorized image. The colorization performance of the fusion model has also been compared with an approach where we simply perform a weighted overlay of the foreground and background proposals to obtain the final output. The weighting factor $\alpha$ determines the proportion of foreground colorized output in the final image. We have used the COCO Validation split  dataset to report the FID obtained by each approach and the results are discussed in Table \ref{alpha}.  
\begin{table}[h]
\centering
\begin{tabular}{ l|lll|l } 
\hline
$\alpha$         & 0.00 & 0.50 & 1.00 & Fusion\\
FID $(\downarrow)$   & 22.59 & 19.45 & 21.37 & 17.60\\
\hline
\end{tabular}
\caption{Fusion Model Ablation study: Comparison with weighted overlay of foreground and background proposals.}
\label{alpha}
\end{table}

\section{Runtime Analysis}
\label{runtime}
The weight parameter file size of our complete colorization framework is less than 250 MB which is significantly lesser than the available state-of-the-art learning-based models for grayscale image colorization. We perform extensive runtime analysis of our approach by using NVIDIA GTX 1080 GPU with 8GB of graphics memory.
\begin{figure}[h]
\centering
\includegraphics[width=0.49\textwidth]{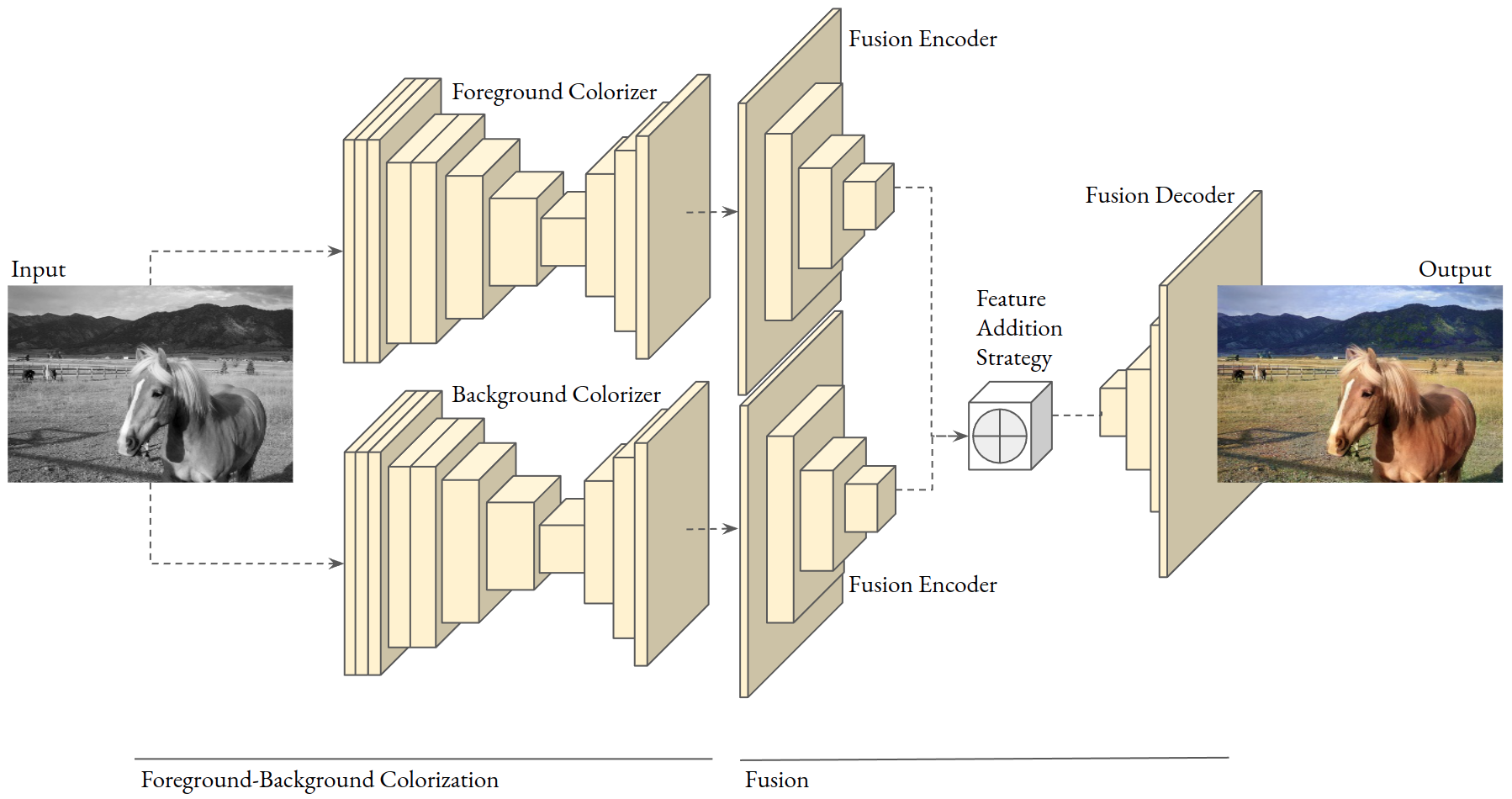}
\caption{Testing Pipeline}
\label{testing}
\end{figure}
Since our method consists of two stages, namely, the parallel colorization step and the feature-based adaptive fusion stage, we report the average inference time computed using a thousand randomly selected images from the COCO Validation Split dataset for each stage. It is to be noted that the object detector is not required in the testing pipeline since the model performs colorization based on full image features only. The parallel colorization stage takes an average time of 0.015 seconds to execute and the fusion pipeline reports an average of 0.009 seconds. Therefore the complete testing pipeline represented in Figure \ref{testing} takes an average time of around 0.024 seconds to produce final colorized outputs of resolution 256x256.

\section{Discussion} 
\label{discussion}
\begin{figure}[h]
\centering
\includegraphics[width=0.48\textwidth]{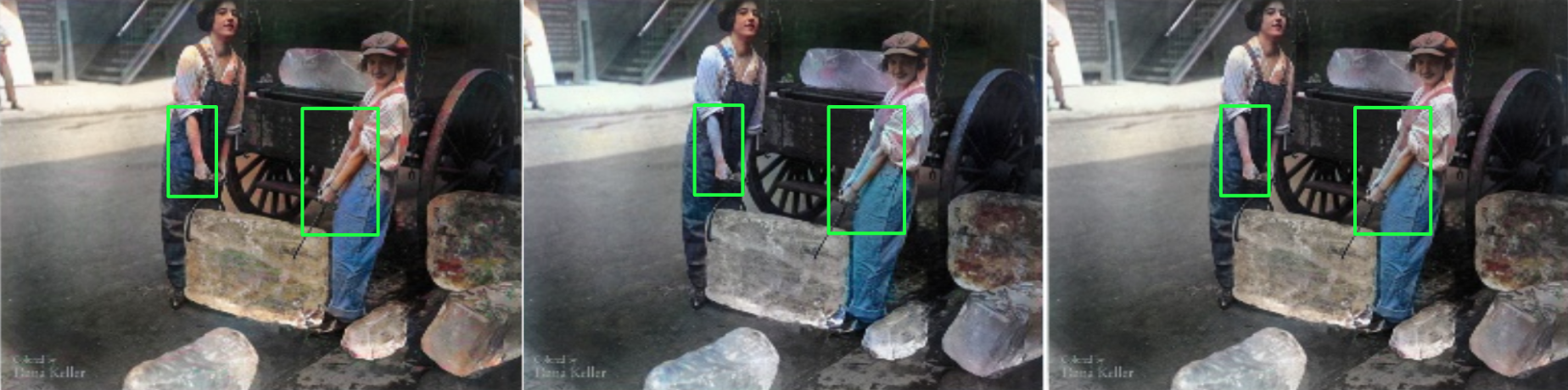}
\caption{Left image is the object colorized output, middle image is the background colorized output and the right image is the fused output.}
\label{fail}
\end{figure}

The problem of image colorization and addressing it using a learning-based algorithm provides important insight toward designing color-predictive models that understand the context of the scene. Earlier works on image colorization have identified the problem of multi-modality but have failed to evaluate their methods using metrics that account for multi-modality. We observed that preserving the semantic information and maintaining the balance between foreground and background color appearance using a single network is hard to achieve even in the state-of-the-art. This became a major reason we came up with an approach that leverages parallel training to preserve the foreground-background information. We separately modified the networks and trained them parallelly. We successfully merged their outputs to retain the best features to produce a balanced image.
The image shown in Figure \ref{fail} is a case where the foreground colorizer has performed well and assigned proper colors to the output, but the background colorizer has left a lot of blue artifacts that have been retained even after the fusion. The probable solution to this could be training the fusion model using the derived recolorized images from the colorizer pipelines.

\section{Conclusion}
\label{conclusion}
When the problem of image colorization is subdivided into two tasks of foreground and background colorization, we observed that a significant improvement in visual performance can be obtained with a substantially lesser amount of training data. Employing multiple models helps address the diverse data modes that alleviate the problem of the desaturated background while colorizing the foreground without leaving color-bleeding artifacts. Parallel GANs adversarially trained to produce the foreground and background colorizer models act as compensators for each other so the fusion network can efficiently combine the specialization areas of both the colorizer models thus obtaining a balanced final image. Since our main focus was to produce colorized images that look realistic, we performed a blind human evaluation of our results based on the visual realism of the colorized images. The testing pipeline instantly produces colorized images of resolution 256x256. After an extensive performance evaluation against existing state-of-the-art methods, we showed that our results achieved quantitative scores comparable to the state-of-the-art with significantly less training data.

\section*{Acknowledgements} 
This work has been performed using the GPU resources provided by the AI Computation Facility of Intelligent Systems Group. The authors would like to thank the Knowledge Resource Centre at Central Electronics Engineering Research Institute (CSIR-CEERI), India for providing the required information resources for their research.  

\section*{Compliance with ethical standards}
\textbf{Conflict of interest} The authors declare that they have no conflict of interest.
 
\bibliographystyle{IEEEtran}
\bibliography {doc.bib}

\vfill

\end{document}